\pdfoutput=1

\documentclass[11pt]{article}

\usepackage{EACL2023}
\usepackage{adjustbox}
\usepackage{times}
\usepackage{soul}
\usepackage{latexsym}
\usepackage{booktabs,siunitx,rotating}
\usepackage{multirow}
\usepackage{enumitem}
\usepackage{microtype}
\usepackage{graphicx}
\usepackage{subfigure}
\usepackage{booktabs} 
\usepackage[T1]{fontenc}

\usepackage[utf8]{inputenc}

\usepackage{microtype}

\newcommand{\Ni}{({\em i})~}
\newcommand{\Nii}{({\em ii})~}

\title{Are the Best Multilingual Document Embeddings \\ simply Based on Sentence Embeddings?}

\author{Sonal Sannigrahi,\textsuperscript{{\normalfont 1}} {\bf Josef van Genabith,}\textsuperscript{1,2}  {\bf Cristina Espa\~{n}a-Bonet}\textsuperscript {{\normalfont 2}} \\  
  \textsuperscript{1}Saarland University, Saarland Informatics Campus, Germany \\
  \textsuperscript{2}German Research Center for Artificial Intelligence (DFKI) 
  \\
 {\tt sosa00001@stud.uni-saarland.de}\\
  {\tt \{cristinae, Josef.Van\_Genabith\}@dfki.de }\\
  }

\begin{document}
\maketitle
\begin{abstract}
Dense vector representations for textual data are crucial in modern NLP. Word embeddings and sentence embeddings estimated from raw texts are key in achieving state-of-the-art results in various tasks requiring semantic understanding. However, obtaining embeddings at the document level is challenging due to computational requirements and lack of appropriate data. Instead, most approaches fall back on computing document embeddings based on sentence representations. Although there exist architectures and models to encode documents fully, they are in general limited to English and few other high-resourced languages. In this work, we provide a systematic comparison of methods to produce document-level representations from sentences based on LASER, LaBSE, and Sentence BERT pre-trained multilingual models. We compare input token number truncation, sentence averaging as well as some simple windowing and in some cases new augmented and learnable approaches, on 3 multi- and cross-lingual tasks in 8 languages belonging to 3 different language families.
Our task-based extrinsic evaluations show that, independently of the language, a clever combination of sentence embeddings is usually better than encoding the full document as a single unit, even when this is possible. We demonstrate that while a simple sentence average results in a strong baseline for classification tasks,  more complex combinations are necessary for semantic tasks. 
Our code is publicly available.%
\footnote{\url{https://github.com/sonalsannigrahi/Document_Embeddings}}
\end{abstract}

\section{Introduction}
\label{s:intro}


Semantic representations, especially embeddings, are crucial for natural language processing (NLP). In fact, the field has exploded since the success of dense word embeddings \cite{mikolov2013distributed}. For some tasks like finding semantic or syntactic relations among words, high quality word embeddings are enough. Other tasks, like question classification or paraphrase detection, benefit from sentence embeddings. Finally, lots of tasks deal with documents: summarisation, document classification, question answering, etc. Document representations are difficult to be learned, especially multilingually, given the amount of available training data and the length of each training instance. 

For these reasons, document embeddings usually resort to sentence embeddings. Since some of the state-of-the-art techniques for language modelling and sentence embeddings are based on self-attention architectures such as BERT~\cite{devlin2019bert}, and self-attention scales quadratically 
with the input length, one cannot afford arbitrarily long inputs.  Training is usually constrained to input fragments up to 512 tokens (subunits). This limit goes well beyond an average sentence length and can cover several paragraphs. However, full documents can be significantly longer. The average length of a Wikipedia article in English is 647 words (not subunits) for example,%
\footnote{\url{https://en.wikipedia.org/wiki/Wikipedia:Size_of_Wikipedia} \\ Consulted on Feb. 2023.} and the average for two of the tasks that we consider in this work, document alignment and ICD code classification, is around 800 words, with documents up to 40k words.

In order to be able to process long inputs, more efficient architectures such as Linformer~\citep{linformer}, Big Bird~\citep{zaheer2020bigbird} or Longformer~\citep{longformer} implement sparse attention mechanisms that scale linearly instead of quadratically. These architectures accept at least 4096 input tokens. With this length, one can embed most Wikipedia articles, news articles, medical records, etc. These architectures are available as pre-trained models in English%
\footnote{\url{https://huggingface.co}}
and can be fine-tuned for NLP tasks such as document classification, question answering or summarisation. However, multilingual or non-English versions are rare. For most languages, it is not just a matter of training a model from scratch, but the amount of documents is just not enough to train high quality models.



LASER \citep{artetxe-schwenk-2019-massively, heffernan2022bitext}, Sentence BERT ~\citep{reimers-gurevych-2019-sentence,reimers-gurevych-2020-making} and LaBSE \citep{feng-etal-2022-language} are representative and state-of-the-art models which largely adapt language models to be used as task-independent sentence representations. These models are  available as pre-trained models and, contrary to the long sequence models introduced before, they are multilingual.  LASER, which is not transformer-based, allows longer inputs.

These observations explain why the two main approaches to obtain multilingual (or non-English) document embeddings are simply \Ni truncating the input to 512 tokens and feeding it into a sentence-level encoder or \Nii splitting the document in shorter fragments and then combine their embeddings. There are few works that do a systematic comparison among methods. \citet{park-etal-2022-efficient} perform a systematic study for document classification in English and found that the most sophisticated models such as Longformer do not always improve on a baseline that truncates the input to fit it into a fine-tuned BERT. The results mostly depend on how the information is distributed along a document and therefore varies from dataset to dataset. 

In this work we explore multilingual document-level embeddings in three tasks in detail: \textit{document alignment}, a bilingual semantic task; \textit{ICD code (multi-label) classification} in 2 languages; and \textit{cross-lingual document classification} in 8 languages. We compare input token number truncation, sentence averaging
as well as some simple windowing and
in some cases new augmented and learnable approaches. Our results show that a simple sentence average is a very strong baseline, even better than considering the whole document as a single unit, but that positional information is needed when the distribution of information across a document is not uniform.


\section{Related Work}
\label{sec:sota}

Word embeddings have been exceptionally successful in many NLP applications \cite{mikolov2013distributed, pennington2014glove, fasttext}. Subsequent works developed methods to learn continuous vector representations for longer sequences  such as sentences or even documents. Skip-thought embeddings~\cite{kiros2015skip} train an encoder--decoder architecture to predict surrounding sentences. \citet{conneau-EtAl:2017:EMNLP2017} showed that the task on which sentence representations are learnt significantly impacts their quality. InferSent \cite{conneau-EtAl:2017:EMNLP2017}, a Siamese BiLSTM network with max pooling, and Universal Sentence Encoder \cite{46808}, a transformer-based network, are trained over the SNLI dataset which is suitable for learning semantic representations \cite{snli:emnlp2015}. 

These methods primarily work on a single language but as multilingual representations have attracted more interest, sentence-level embeddings have been extended to obtain a wider language coverage. 
\newcite{artetxe-schwenk-2019-massively} (LASER) learn joint multilingual sentence representations for 93 languages based on a single BiLSTM encoder with a shared BPE vocabulary trained on publicly available parallel corpora. However, this architecture was shown to underperform in high-resource scenarios \cite{feng-etal-2022-language}. LASER is especially interesting for our work as, being LSTM-based, it does not have the 512-length constraint. 
\citet{litransformer} introduce T-LASER, which is a version of LASER that uses a transformer encoder in place of the original bidirectional LSTM. However, this model was tested only on the Multilingual Document Classification (MLDoc) corpus \cite{schwenk-li-2018-corpus}, which does not have significantly long documents.
Similarly, \newcite{reimers-gurevych-2019-sentence} (sBERT in the following) extended a transformer-encoder architecture, BERT, by using a Siamese network with cosine similarity for contrastive learning in order to derive semantically meaningful sentence representations. More recently, \newcite{feng-etal-2022-language} (LaBSE) explored cross-lingual sentence embeddings with BERT by introducing a pre-trained multilingual language model component and show that on several benchmarks, their method outperforms many state-of-the-art embeddings such as LASER. 

While sentence-level representations have been widely explored in literature, document-level representations are less well-explored. The earliest approaches in learning document-level vector representations included an extension of the Word2Vec algorithm named Doc2Vec \cite{le2014distributed} with two variants proposed, a bag-of-words and a skip-gram based model. However, while these methods worked well at the word-level, the document-level counterpart led to issues in scaling due to large vocabulary sizes \cite{lau2016empirical}. Due to these limitations, further works have attempted to improve the computational bottlenecks involved with training on long sequences such as documents. Linformer \citep{linformer} is a transformer-based architecture with linear complexity due to a sparse self-attention mechanism making it significantly more memory- and time-efficient in comparison with the original transformer \cite{vaswani2017attention}. Works such as Big Bird~\citep{zaheer2020bigbird} and Longformer~\citep{longformer} introduced a sparse attention mechanism and localised global attention respectively. BigBird is able to handle sequences of up to 4,096 tokens and Longformer scales linearly with the sequence length, with experiments on sequences of length upto 32,256. To the best of our knowledge, to date not much has been done to extend them beyond English.  \citet{longformer-base-4096-chinese} and \citet{mromero2022longformer-base-4096-spanish} made available Chinese and Spanish Longformer models, respectively, while \citet{Sagen1545786} trained a multilingual version starting from a RoBERTa checkpoint and not from scratch. We use Longformer as a comparison system in our experiments but we do not consider the multilingual model given that multilinguality was achieved by fine-tuning on question answering data and we do not explore this task.

\section{Sentence Embeddings}
\label{s:sentenceEmb}

We use three multilingual sentence-level embedding models that cover different languages, architectures and learning objectives:

\paragraph{LASER} \citep{schwenk-douze-2017-learning, artetxe-schwenk-2019-massively} uses max-pooling over the output of a stacked BiLSTM-encoder. The encoder is extracted from an encoder--decoder machine translation setup trained on parallel corpora over 93 languages. Since it is not based on transformers but on LSTMs, the maximum number of input tokens can in principle be arbitrary and is set to 12,000.

\paragraph{LaBSE} \newcite{feng-etal-2022-language} train a multilingual BERT-like model with a masked LM and translation LM objective functions.
A dual-encoder transformer is initialised with the model and fine-tuned on a translation ranking task. The final model covers 109 languages. The maximum number of input tokens is 512.

\paragraph{sBERT} \newcite{reimers-gurevych-2019-sentence} use the output of BERT-base with mean pooling to create a fixed-size sentence representation. A Siamese-BERT architecture trained on NLI is used to obtain the final sentence-embedding model. The maximum number of input tokens is 512, with a default value of 128. We use the multilingual version \cite{reimers-gurevych-2020-making}.

\section{Document Embeddings}
\label{sec:docEmb}


We divide our approaches to build document embeddings into three families: in \textit{(i) Document Excerpts,} we feed token sequences as they are directly into LASER, LaBSE and sBERT to obtain a document-level representation, in \textit{(ii) Sentence Weighting Schemes,} we divide documents into sentences represented using base sentence embeddings and then explore different combination and weight strategies to obtain document embeddings, in \textit{(iii) Windowing Approaches,} we study different distributions to learn document-level positional and semantic information. 

\paragraph{(i) Document Excerpts}  ~
\paragraph{\textit{All Tokens:}} The full document is fed into the system (no truncation). We explore this option only with LASER which does not have the 510-token-length restriction%
\footnote{That is the maximum length of tokens accepted by transformer-style embedding models, 512 without the \texttt{[CLS]} and \texttt{[SEP]} tokens.} 
and when possible (English, Spanish and Chinese) with Longformer.

\paragraph{\textit{Top-N Tokens:}} The document is truncated to the first $n=510$ tokens.
    
\paragraph{\textit{Bottom-N Tokens:}} The last $n=510$ tokens are fed into the system. 
    
\paragraph{\textit{Top-N + Bottom-M Tokens:}} We select $N=128$ and $M=382$ to use the first $N$ and last $M$ tokens of the documents. These values are based on empirical explorations by \newcite{sun2019fine}.

\medskip
\paragraph{(ii) Sentence Weighting Schemes} 
\paragraph{\textit{Sentence Average:}} Each base sentence embedding (obtained with LASER, LaBSE or SBERT) is given a uniform weight. This computes the vanilla average embedding vector of all sentences in the document.
\paragraph{\textit{Top/Bottom-Half Average:}} Only the top (bottom) half of the sentences in the document are considered for averaging.  
    \paragraph{\textit{TF-IDF Weights:}} We compute TF-IDF scores for all terms in a document, and average their values at sentence level. The base sentence embeddings (LASER, LaBSE, SBERT) are then weighted by the normalised value of the TF-IDF averages. Following \citet{buck-koehn-2016-quick}, we use different TF-IDF computations based on variations of term frequency $tf$ and inverse document frequency $idf$ definitions. For words $w$ in a document $d$ belonging to a collection $D$ we report results using:
    \begin{equation}
         tf_2(w, d) = \textrm{freq}(w, d)
         \label{eq:tf2}
    \end{equation}
    \begin{equation}
        tf_4(w, d) =0.4 + 0.6 \frac{\textrm{freq}(w, d)}{\textrm{max}_{\tilde{w}} \textrm{freq}(\tilde{w}, d)} 
         \label{eq:tf4}
    \end{equation}
   \begin{equation}
       idf_4(w, d) = \textrm{log}(1+ \frac{|D|}{\textit{df}(w, |D|)}),
         \label{eq:idf4}
   \end{equation}

\noindent
with $df(w, D) = |\{ d \in D | w \in d\}|$, and 
\begin{equation}
       tf_iidf_j(S_k) = \frac{\sum_{w \in S_k} tf_i(w, d)idf_j(w, d)}{\#w_k},
       \label{eq:tfidf}
   \end{equation}
where  $S_k$ is a sentence in a given document $d$, and $\#w_k$ is the number of words in sentence $S_k$.

\medskip
The weights of these models are fixed for the static tasks and used as initialisation when training a classifier.

\bigskip
\paragraph{(iii) Windowing Approaches} 

\paragraph{\textit{TK-PERT:}}
\citet{thompson-koehn-2020-exploiting} introduced a windowing approach that weights the contribution of each sentence according to the modified PERT function \cite{vose2008risk} and a down-weighting function for boilerplate text. The latter was introduced to deal with webpages but it can be ignored for other types of documents. The smoothed overlapping windowing functions based on a cache of the PERT distribution (PERT-cache) encode fine-grained positional information into the resultant document vector.

A document with $N$ sentences $S_{i | i \in \{0, \dots , N-1 \}}$ is split uniformly into $J$ parts and the final representation $D$ for a document is given by a concatenation of normalised position-weighted (via PERT) sub-vectors where each sub-vector $D_j$ is
\begin{equation}
    D_j = \sum_{n=0}^{N-1}\textrm{emb}(S_n) P_j(n) B(S_n),
    \label{eq:tkpert}
\end{equation}
\noindent  emb is the (LASER, LaBSE, SBERT) embedding of sentence $n$, $P$ is the modified PERT function for part $j$ and $B$ is a boilerplate function if there is one. In cases when no boilerplate text is present, we set it to 1.

Following \citet{thompson-koehn-2020-exploiting} setting for the modified PERT distribution, we use $J=16$ and set its shape parameter to $\gamma=20$.


\begin{figure}[t]
    \hspace*{-0.5em}
    \includegraphics[width=0.5\textwidth]{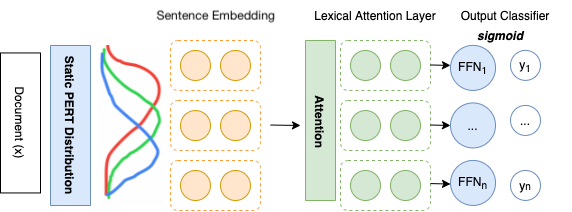}
    \caption{ATT-PERT model for classification. A static modified PERT distribution is used to extend the sentence embeddings to documents. Afterwards, an attention-weighted classifier is learnt.}
    \label{fig:maps}
\end{figure}

\paragraph{\textit{TF-PERT:}} is a new extension of TK-PERT to further incorporate semantics. PERT focuses on positional information encoded in the document while TF-IDF focuses on the semantic information, therefore a combined metric would likely be able to consider both features. We combine the two contributions with a multiplication at sentence level:
\begin{equation}
    D_j = \sum_{n=0}^{N-1}\textrm{emb}(S_n) P_j(n) B(S_n) tfidf(S_n),
\end{equation}
where we use the same notation as in Eqs.~\ref{eq:tfidf} and \ref{eq:tkpert}.

\paragraph{\textit{ATT-PERT:}} is a new extension of TK-PERT to further incorporate a global learnable attention. 
Figure \ref{fig:maps} illustrates the basic architecture. The PERT distribution encodes global positional information of the document.  By adding an attention layer over it, we introduce a \textit{global attention} that weights the different parts of the document and that is combined with the standard \textit{local attention} at word level performed by the sentence encoder. Mathematically, 
\begin{equation}
    D_j =  \sum_{n=0}^{N-1}\textrm{emb}(S_n) P_j(n) a_j(n),
\end{equation}
\noindent where $S_n$ refers to the sentence embedding that has been trained for a classification task and $a_j(n)$ is the respective global attention weight.


In TK-PERT, the static PERT distribution is multiplied by the fine-tuned sentence embeddings. In contrast, in ATT-PERT, the distribution is multiplied with the embeddings prior to training a classifier without freezing the embedding layer, as this allows the positional weights in the PERT distribution to be trained for the specific task.

\paragraph{\textit{ATT-TF-PERT:}} is a new extension of TF-PERT to further incorporate a global learnable attention as in ATT-PERT. 
In this configuration, we learn combined TF-IDF-PERT weighted embeddings whose attention weights are further updated while training the classifier. We use the same \textit{global attention} $a_j(n)$ as in ATT-PERT, however here it is multiplied with both the TF-IDF weight of the sentence $tfidf_j(w,S_n)$ as computed in the TF-IDF set up and the PERT distribution  $P_j(n)$ as in TK-PERT: 
\begin{equation}
 D_j = \sum_{n=0}^{N-1}\textrm{emb}(S_n) P_j(n) a_j(n) tfidf(S_n).
\end{equation}






\begin{table}[t]
    \centering
    \small
\resizebox{0.49\textwidth}{!}{%
\begin{tabular}{lrrrr}
    \toprule
        & \multicolumn{2}{c}{\# Documents} & \multicolumn{2}{c}{Length} \\
  & Train & Test &  Avg. & Max. \\
    \midrule
    \multicolumn{5}{c}{\it Document Alignment, WMT2016}\\
    \cmidrule(lr){1-5}
         English & 349k & 682k & 737 & 43.3k\\
         French & 225k & 522k & 842 & 45.2k \\
         Web Domains & 49 & 203 & - & -\\ 
         Gold Pairs & 1624 & 2402 & -& -\\
    \midrule
    \multicolumn{5}{c}{\it Multi-label Classification, ICD Code Classification}\\
    \cmidrule(lr){1-5}
        Spanish & 1001 & 1600 & 792 & 4352 \\
        German & 8385 & 407 & 876 &  2249\\
    \midrule
    \multicolumn{5}{c}{\it Document Classification, MLDoc} \\
    \cmidrule(lr){1-5}
         English & 10k & 4000 & 275 & 576\\
         German & 10k & 4000 & 342 & 675 \\
         French & 10k & 4000 & 445 & 782\\
         Italian & 10k & 4000 & 376 & 765\\
         Spanish & 9458 & 4000 & 354 & 778\\
         Japanese & 10k & 4000 & 327 & 897\\
         Russian & 5216 & 4000 & 235 & 967\\
         Chinese & 10k & 4000 & 562 & 983\\
    \bottomrule
    \end{tabular}
    }
  \caption{Number of documents and average tokenised document length in sentencepiece units (prior to boilerplate downweighting for Document Alignment) for the three tasks used in the experiments.}
    \label{tab:data}
\end{table}

\section{Evaluation Tasks}
\label{s:tasks}

We apply the different configurations discussed above across the following tasks:


\paragraph{Bilingual Document Alignment} aims at aligning documents from two collections in language L1 and language L2 according to whether they are parallel or comparable. In our experiments, we use the data given for the WMT 2016 Shared Task on Bilingual Document Alignment to align French web pages to English web pages for a given crawled webdomain~\cite{buck-koehn-2016-findings}. In these experiments we do not perform any learning using the training data, but just estimate document-level semantic similarity between the pairs of documents in the test set. To compute this, we find the top $K$=$32$ candidate translations using approximate nearest neighbor search via FAISS%
\footnote{\url{https://github.com/facebookresearch/faiss}}
as in \cite{buck-koehn-2016-findings}. We use cosine similarity to quantify semantic similarity on the document embeddings.


\paragraph{Multi-label ICD Code Classification} aims at assigning one or more ICD-10 codes to medical-domain texts (electronic health records). Here there can be an arbitrary number of ICD-10 codes assigned to the input text. In particular, out of all the possible ICD-10 Codes, 4 account for more than 90\% of the documents, making this an imbalanced classification task and leading to the 'tail end problem' \cite{chapman2020automatic}. 
We use the CLEF eHealth 2019 task for German non-technical summaries \cite{Neves_Butzke_Dorendahl_Leich_Grune_Schonfelder_2019} and CANTEMIST-CODING \cite{miranda2020named} for Spanish electronic health records. Here, we learn a weighted-attention classifier layer \cite{lee2022weight} 
on top of the base document embeddings
consisting of a feed-forward neural network with a single hidden layer of 10 units. 


\paragraph{Cross-lingual Document Classification} aims at classifying documents in a set of predefined categories in a language (usually English) and then transfer the model to unseen languages.
We use the MLDoc dataset
for this purpose \cite{schwenk-li-2018-corpus}.
The corpus contains 1,000 development documents and 4,000 test documents in eight languages (English, German, French, Italian, Spanish, Japanese, Russian and Chinese), divided in four different genres with uniform class priors. 
For zero-shot transfer, we train a classifier on top of the multilingual document representations estimated as described in Section~\ref{sec:docEmb} by using only the English training data and the hyperparameters optimised in \newcite{artetxe-schwenk-2019-massively}. Similar to the previous classification task, we use a feed-forward neural network with one hidden layer with 10 units. We use this classifier on top on the multilingual embeddings to evaluate the system on the remaining languages.



\bigskip
Table \ref{tab:data} shows the statistics for the datasets used in the three tasks as well as an average length of training instances in terms of sentencepiece tokens.%
\footnote{\url{https://github.com/google/sentencepiece}} The average document length in the document alignment and ICD code classification tasks is larger than 512 tokens, making the usage of sentence embeddings alone insufficient.
This is not the case for document classification, but we still consider it in order to compare the different approaches and add a highly multilingual setting.


\section{Results and Discussion}
\label{sec:results}

\citet{thompson-koehn-2020-exploiting} empirically obtained the best trade-off between accuracy and inference time when using PCA-reduced sentence embeddings of 128 dimensions in the bilingual document alignment task. We performed equivalent experiments with 128 and 256 dimensions for selected configurations in the three tasks and confirmed the trend. As we obtained no major gains in using more dimensions, we report all the results for the three tasks with 128-dimensional sentence embeddings.  

We report confidence intervals at 95\% confidence level using bootstrap resampling with 1000 samples for document alignment, 500 samples for ICD code classification and 1000 samples for document classification.

\begin{table}[t]
\setlength\tabcolsep{4pt}
 \resizebox{\columnwidth}{!}{%
\small
 \begin{tabular}{lrrr}
    \toprule
    & \textbf{LASER} & \textbf{LaBSE} & \textbf{sBERT}\\
      \midrule
    All tokens         & \textbf{81.2$^{+0.3}_{-0.4}$} & --- & ---\\
    Top-510 tokens     & 70.8$^{+0.2}_{-0.3}$ & 71.2$^{+0.5}_{-0.4}$ & 72.3$^{+0.2}_{-0.4}$\\
    Bottom-510 tokens  & 65.8$^{+0.5}_{-0.3}$ & 66.3$^{+0.7}_{-0.8}$ & 67.1$^{+0.6}_{-0.7}$\\
    Top-128 + Bot-312  & 75.3$^{+0.5}_{-0.5}$ & \textbf{76.1$^{+0.3}_{-0.5}$} & \textbf{74.2$^{+0.3}_{-0.3}$}\\
    \midrule
    Sentence Average & 81.8$^{+0.7}_{-0.5}$ & 83.4$^{+0.6}_{-0.5}$ & 82.3$^{+0.4}_{-0.6}$ \\
    Top-Half Avg.    & 82.2$^{+0.3}_{-0.5}$ & 81.3$^{+0.6}_{-0.8}$ & 81.7$^{+0.7}_{-0.6}$ \\
    Bottom-Half Avg. & 67.8$^{+0.8}_{-0.7}$ & 66.5$^{+0.4}_{-0.3}$ & 65.3$^{+0.5}_{-0.4}$ \\
    TF-IDF Weighted  &  \\
    ~~~~\textit{$tf_2-idf_4$} & 80.2$^{+0.7}_{-0.4}$ & 80.5$^{+0.7}_{-0.6}$ & 79.3$^{+0.2}_{-0.4}$\\
    ~~~~\textit{$tf_4-idf_4$} & \textbf{86.3$^{+0.5}_{-0.4}$} & \textbf{87.2$^{+0.3}_{-0.4}$} & \textbf{85.4$^{+0.6}_{-0.5}$}\\
    \midrule
    TK-PERT (Euclidean)  & 93.2$^{+0.7}_{-0.8}$ & 93.5$^{+0.6}_{-0.5}$ & 92.8$^{+0.5}_{-0.4}$\\
    TK-PERT (cosine)     & \textbf{96.4$^{+0.6}_{-0.5}$} & \textbf{94.2$^{+0.5}_{-0.4}$} & \textbf{95.3$^{+0.8}_{-0.9}$} \\
    TF-PERT (cosine) & 93.4$^{+0.5}_{-0.3}$ & 92.5$^{+0.3}_{-0.4}$ & 93.1$^{+0.4}_{-0.4}$ \\ 
    \bottomrule
  \end{tabular}
}
\caption{Document recall on WMT-16 Shared Task on English--French document alignment. Best score for each family is in bold.}
  \label{tab:baselines}
\end{table}


\begin{table*}[t]
\centering
\small
\setlength\tabcolsep{8pt}
\begin{tabular}{lrrrrrr}
    \toprule
    & \multicolumn{2}{c}{\bf LASER} & \multicolumn{2}{c}{\bf LaBSE} & \multicolumn{2}{c}{\bf sBERT} \\
    & \multicolumn{1}{c}{\bf de} & \multicolumn{1}{c}{\bf es} & \multicolumn{1}{c}{\bf de} & \multicolumn{1}{c}{\bf es}  & \multicolumn{1}{c}{\bf de} & \multicolumn{1}{c}{\bf es} \\
      \midrule
    All tokens        & \textbf{73.1$^{+0.5}_{-0.6}$} & \textit{18.4$^{+0.4}_{-0.3}$} & -- &-- & -- &--\\
    Top-510 tokens    & 65.6$^{+0.8}_{-0.7}$ & 16.5$^{+0.5}_{-0.8}$ & 68.2$^{+0.5}_{-0.5}$ & \textit{19.2$^{+0.6}_{-0.4}$}  & 63.2$^{+0.7}_{-0.8}$ & \textit{18.3$^{+0.5}_{-0.6}$} \\
    Bottom-510 tokens & 67.8$^{+0.4}_{-0.9}$ & 17.5$^{+0.4}_{-0.9}$ & 66.7$^{+0.8}_{-0.6}$ & 17.4$^{+0.7}_{-0.6}$  & 61.5$^{+0.8}_{-0.6}$ & 16.8$^{+0.5}_{-0.7}$  \\
    Top-128 + Bot-312 & 66.4$^{+0.8}_{-0.6}$ & 17.2$^{+0.8}_{-0.7}$ & \textit{69.1$^{+0.7}_{-0.9}$} & 18.7$^{+0.7}_{-0.8}$  & \textit{64.8$^{+0.7}_{-0.5}$} & 17.5$^{+0.6}_{-0.8}$  \\
    \midrule
    Sentence Average & \textit{72.1$^{+0.9}_{-0.8}$} & 17.0$^{+0.7}_{-0.6}$ &\textbf{74.5$^{+0.8}_{-0.9}$} & \textit{24.2$^{+0.8}_{-0.6}$}  & \textbf{68.9$^{+0.7}_{-0.6}$} & \textit{20.3$^{+0.8}_{-0.4}$}  \\
    Top-Half Avg.    & 68.4$^{+0.7}_{-0.9}$ & 16.5$^{+0.5}_{-0.6}$ & 68.3$^{+0.4}_{-0.8}$ & 18.9$^{+0.5}_{-0.5}$  & 61.5$^{+0.7}_{-0.6}$ & 16.4$^{+0.8}_{-0.6}$\\
    Bottom-Half Avg. & 63.1$^{+0.7}_{-0.6}$& 15.8$^{+0.8}_{-0.7}$ & 67.4$^{+0.8}_{-0.9}$ & 15.2$^{+0.6}_{-0.7}$  & 58.6$^{+0.9}_{-0.8}$ & 17.9$^{+0.7}_{-0.6}$\\
    TF-IDF Weighted  & 65.3$^{+0.5}_{-0.4}$ & \textit{17.2$^{+0.7}_{-0.8}$} & 68.2$^{+0.9}_{-1.0}$& 19.2$^{+0.9}_{-0.7}$  & 63.2$^{+0.6}_{-0.8}$ & 18.3$^{+0.7}_{-0.6}$ \\
    \midrule
    TK-PERT  & 68.2$^{+0.8}_{-0.6}$ & 22.1$^{+0.7}_{-0.4}$ & 70.1$^{+0.8}_{-0.7}$ & 20.1$^{+0.7}_{-0.4}$ & 65.2$^{+0.8}_{-0.6}$ & 19.5$^{+0.7}_{-0.8}$  \\
    TF-PERT  & 68.5$^{+0.4}_{-0.3}$ & 23.4$^{+0.6}_{-0.6}$ & 68.6$^{+0.3}_{-0.7}$ & 21.3$^{+0.5}_{-0.4}$ & 65.4$^{+0.6}_{-0.7}$ & 18.7$^{+0.4}_{-0.3}$  \\    
    ATT-PERT & \textit{70.7$^{+0.7}_{-0.9}$} & \textbf{32.2$^{+0.7}_{-0.4}$} & 72.1$^{+0.8}_{-0.6}$& \textbf{30.1$^{+0.7}_{-0.8}$}  &  \textit{66.3$^{+1.1}_{-1.3}$} & \textbf{27.4$^{+0.8}_{-0.7}$}\\    
    ATT-TF-PERT & 70.3$^{+0.5}_{-0.4}$ & 31.4$^{+0.8}_{-0.7}$ & \textit{73.2$^{+0.4}_{-0.8}$} & 29.7$^{+0.6}_{-0.5}$ &  66.1$^{+0.9}_{-0.8}$ & 27.1$^{+0.5}_{-0.6}$\\
    \bottomrule
  \end{tabular}
  \caption{F1 scores for the Multi-label ICD code classification task for German (de) and Spanish (es) documents. Best scores are in bold, and best scores per family are in italics. }
  
  \label{tab:icd-classification}
\end{table*}

\begin{figure*}[ht]
    \centering
    \includegraphics[width=0.7\textwidth,angle=270,origin=c]{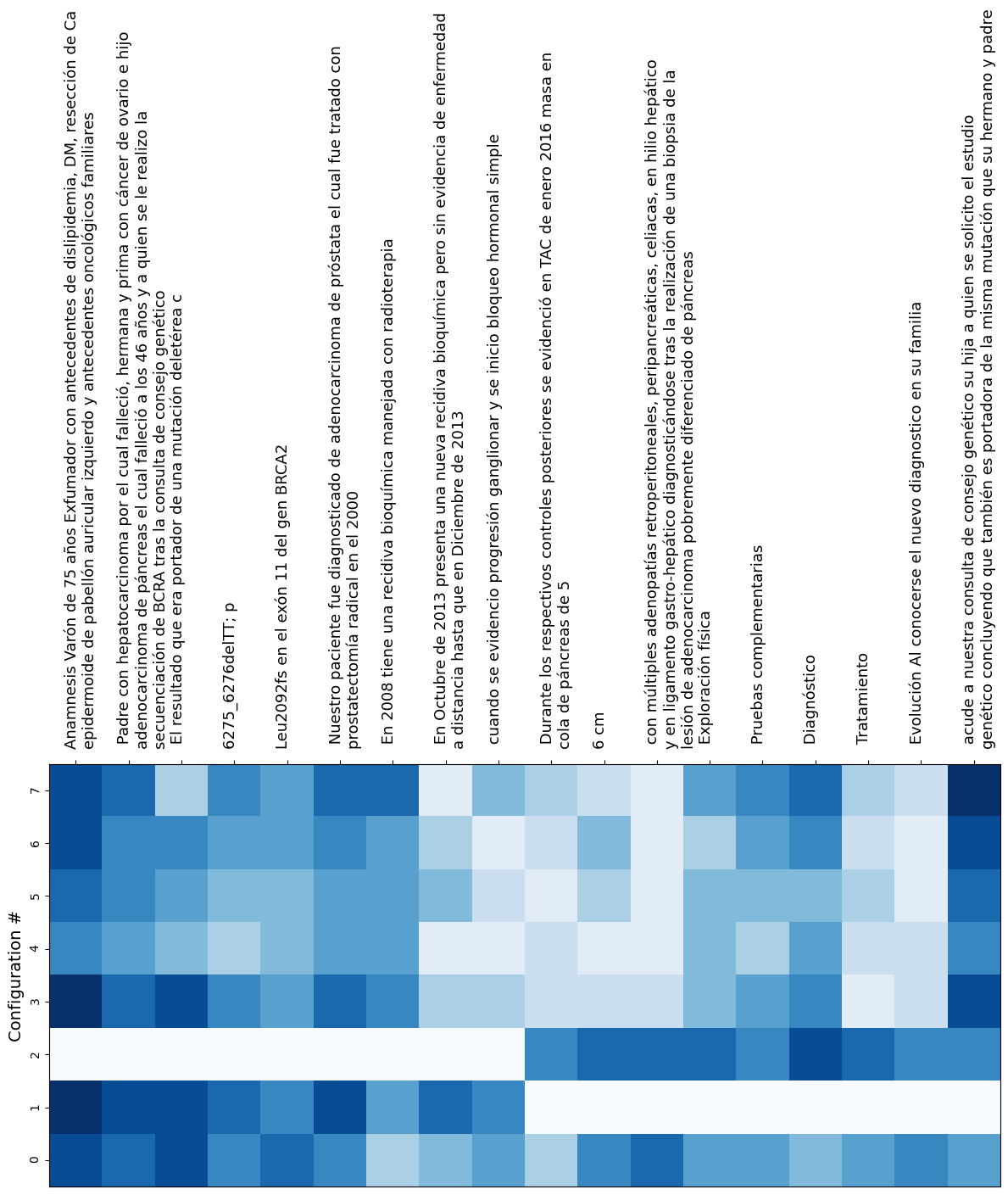}
    \vspace{-3em}
    \caption{Sentence weights for an example document with LASER embeddings and the configurations: 0-Sentence average, 1-Top-Half, 2-Bottom-Half, 3-TF-IDF weighted, 4-TK-PERT, 5-TF-PERT, 6-ATT-PERT, 7-ATT-TF-PERT.}
    \label{fig:scores}
\end{figure*}

\paragraph{Bilingual Document Alignment} quality ranges from 65\% to 96\% recall depending on the document embedding method.
Table \ref{tab:baselines} shows the results obtained for all the configurations considered. 
A simple sentence average achieves a recall around 82\% (depending on the sentence embedding used). When using LASER, the only method that allows the comparison, the recall with sentence average is larger but not statistically significantly over embedding the full document as a single unit (81.8\% vs 81.2\%). Taking a token-based excerpt of the document is 10 percentage points below sentence-averaging the same excerpt. The information in webpages seems to be more densely distributed towards the top of the page. Looking at the top-half versus the bottom-half of the sentences of the webpages, there is a 17\% reduction in the scores obtained.
In these unweighted and average configurations in both the token and sentence-based methods, we do not encode any positional information: sentence order and semantic relevance is not considered in the final document embeddings. However, intuitively, these factors are indicative of each sentence's contribution to the larger document embedding. In order to incorporate semantic relevance into our final embeddings, we consider the weighted average using TF-IDF. 
We explore several TF-IDF forms and obtain a difference of 7\% on average among them. Table~\ref{tab:baselines} shows the 2 most promising ones. With the best option (\textit{$tf_4-idf_4$}), TF-IDF weighting improves between 3 and 5 percentage points with respect to the sentence averaging which uses uniform weights. We use \textit{$tf_4-idf_4$} for the next experiments when required as these formulae empirically performed the best. 
To include sentence order, we use the PERT-window based approach.
TK-PERT outperforms all other methods by a margin of 11.7\%. This result attests the relevance of contextual information, sentence order, and positional importance. Although we find improvements over the baseline models by introducing TF-IDF weights and the PERT distribution, a combination of the two in TF-PERT does not lead to further improvements. 

The other dimension of the study, the particulars of sentence embeddings, is less important to the recall. LASER, LaBSE and sBERT achieve similar results. 
As we are working with French and English documents, both languages being high-resource, all base sentence embeddings are high-quality and therefore they do not impact the final model strongly in a consistent way.

\begin{table*}[t]
    \centering
    \small
    \begin{tabular}{llcccccccc}
\toprule
& &   \multicolumn{8}{c}{\textbf{en $\rightarrow$ xx}}\\
& &  {\bf en} & {\bf de} & {\bf es} & {\bf fr} &{\bf it} & {\bf ja} & {\bf ru} & {\bf zh} \\
\midrule
\multicolumn{2}{l}{Longformer}  & \textbf{\textit{92.3$^{+0.7}_{-0.8}$}} & --& 76.9$^{+0.6}_{-0.7}$ & --& --& --& --& 68.5$^{+0.4}_{-0.5}$\\

\midrule
\multirow{9}{*}{\rotatebox[origin=c]{90}{LASER}} & All tokens & 88.7$^{+1.1}_{-0.8}$ & 83.6$^{+0.5}_{-0.4}$ & 77.4$^{+0.9}_{-0.8}$& 78.1$^{+0.7}_{-0.8}$ & 65.1$^{+0.6}_{-0.7}$& 61.8$^{+0.6}_{-0.4}$ & 66.6$^{+0.5}_{-0.6}$ & 70.1$^{+0.9}_{-0.8}$ \\
&Sentence Average & 89.9$^{+0.9}_{-0.8}$ & 84.8$^{+0.7}_{-0.6}$ & 77.3$^{+0.9}_{-0.7}$ & 77.9$^{+0.5}_{-0.9}$ & \textit{64.9$^{+0.4}_{-0.8}$} & 60.3$^{+0.8}_{-0.7}$ & 67.8$^{+0.8}_{-0.9}$ &\textbf{71.9$^{+0.8}_{-0.7}$} \\
&Top-Half Avg. &  86.4$^{+0.3}_{-0.9}$ & 83.5$^{+0.4}_{-0.5}$ & 75.8$^{+0.9}_{-0.6}$ & 76.2$^{+0.8}_{-0.5}$ & 63.2$^{+0.7}_{-0.9}$ & 56.5$^{+0.7}_{-0.6}$ & 64.1$^{+0.7}_{-0.8}$ & 67.5$^{+0.8}_{-0.7}$\\
&Bottom-Half Avg.  & 83.2$^{+0.4}_{-0.6}$ & 81.4$^{+0.7}_{-0.8}$ & 71.2$^{+0.7}_{-0.6}$ & 70.5$^{+0.8}_{-0.9}$ & 59.2$^{+0.5}_{-0.4}$ & 50.4$^{+0.6}_{-0.7}$ & 56.2$^{+0.6}_{-0.4}$ & 60.3$^{+0.6}_{-0.7}$ \\
&TF-IDF Weighted &  86.3$^{+0.8}_{-0.8}$ & 85.1$^{+0.5}_{-0.8}$ & 75.3$^{+0.7}_{-0.4}$ & 74.1$^{+0.7}_{-0.8}$ & 56.4$^{+0.5}_{-0.7}$ & 61.4$^{+0.6}_{-0.8}$ & 60.2$^{+0.7}_{-0.6}$ & 71.5$^{+0.4}_{-0.5}$\\
&TK-PERT &  89.1$^{+0.4}_{-0.7}$ & 85.2$^{+0.6}_{-0.6}$ & 75.6$^{+0.8}_{-0.7}$ & 78.2$^{+0.8}_{-1.1}$ & 63.6$^{+0.9}_{-0.7}$ & 62.3$^{+0.8}_{-0.4}$ &\textbf{ 67.8$^{+0.6}_{-0.7}$} & 71.1$^{+0.4}_{-0.6}$ \\
&TF-PERT &  88.7$^{+0.6}_{-0.5}$ & 84.8$^{+0.8}_{-0.6}$ & 75.4$^{+0.5}_{-0.4}$& 77.9$^{+0.6}_{-0.4}$ & 61.2$^{+0.9}_{-0.8}$& 61.8$^{+0.3}_{-0.4}$ & 67.2$^{+0.5}_{-0.5}$ & 70.8$^{+0.6}_{-0.5}$\\
&ATT-PERT &  \textit{89.2$^{+0.7}_{-0.8}$} & \textbf{86.2$^{+0.6}_{-0.5}$} & \textbf{77.5$^{+0.8}_{-0.7}$}& \textit{79.1$^{+1.0}_{-0.8}$} & 64.0$^{+0.3}_{-0.9}$ & 62.5$^{+0.6}_{-0.4}$& 66.2$^{+0.8}_{-0.9}$ & 71.3$^{+0.7}_{-0.6}$\\
&ATT-TF-PERT &  88.5$^{+0.6}_{-0.5}$ & 86.0$^{+0.4}_{-0.3}$& 76.7$^{+0.4}_{-0.5}$& 78.9$^{+0.5}_{-0.5}$ & 63.8$^{+0.5}_{-0.6}$& \textbf{62.8$^{+0.5}_{-0.4}$} & 66.5$^{+0.4}_{-0.7}$ & 70.5$^{+0.3}_{-0.5}$\\
\midrule
 \multirow{8}{*}{\rotatebox[origin=c]{90}{LaBSE}} & Sentence Average& \textbf{90.9$^{+0.6}_{-0.3}$} &85.2$^{+0.8}_{-0.7}$ & 75.6$^{+0.5}_{-0.8}$ & \textbf{79.9$^{+0.5}_{-0.3}$} & \textbf{66.9$^{+0.9}_{-0.6}$} & 58.3$^{+0.7}_{-0.6}$ & 65.4$^{+0.5}_{-0.5}$ & 70.1$^{+0.5}_{-0.6}$ \\
& Top-Half Avg. & 86.1$^{+0.7}_{-0.8}$ & 80.5$^{+0.5}_{-0.9}$ & 73.2$^{+0.7}_{-0.8}$ & 76.5$^{+0.9}_{-0.7}$ & 62.5$^{+0.6}_{-0.8}$ & 56.1$^{+0.7}_{-0.6}$ & 61.8$^{+1.0}_{-0.9}$ & 67.3$^{+0.6}_{-0.7}$  \\
&Bottom-Half Avg.  & 85.4$^{+1.2}_{-1.1}$ & 78.7$^{+0.5}_{-0.6}$ & 71.4$^{+0.6}_{-0.7}$ & 73.3$^{+0.8}_{-0.6}$ & 59.6$^{+0.4}_{-0.7}$ & 50.7$^{+0.3}_{-0.8}$ & 58.9$^{+0.7}_{-0.6}$ & 61.4$^{+0.9}_{-1.1}$ \\
&TF-IDF Weighted &  86.2$^{+0.2}_{-0.6}$ & 84.1$^{+0.5}_{-0.4}$& 73.9$^{+0.6}_{-0.3}$ & 77.1$^{+0.3}_{-0.4}$ & 62.6$^{+0.2}_{-0.5}$& 59.3$^{+0.3}_{-0.6}$ & 65.4$^{+0.5}_{-0.4}$ & 68.1$^{+0.8}_{-0.7}$\\
&TK-PERT &   87.1$^{+0.5}_{-0.9}$ & 83.6$^{+0.8}_{-0.6}$ & 75.8$^{+0.5}_{-0.4}$& 79.1$^{+0.7}_{-0.8}$& 62.5$^{+0.3}_{-0.8}$ & 60.0$^{+0.6}_{-0.7}$ & 64.9$^{+0.6}_{-0.4}$& 70.6$^{+0.7}_{-0.6}$\\
&TF-PERT &  86.2$^{+0.5}_{-0.4}$ & 84.7$^{+0.4}_{-0.7}$ & 77.3$^{+0.7}_{-0.6}$& 76.3$^{+0.6}_{-0.5}$ & 62.8$^{+0.5}_{-0.5}$& 61.2$^{+0.7}_{-0.6}$ & 64.5$^{+0.6}_{-0.5}$ & 69.2$^{+0.5}_{-0.6}$\\
&ATT-PERT &  88.9$^{+0.8}_{-0.6}$& 84.3$^{+0.7}_{-0.8}$ & \textit{77.3$^{+0.5}_{-0.5}$} & 79.4$^{+0.7}_{-0.9}$ & 63.8$^{+0.6}_{-0.7}$ & \textit{62.2$^{+0.8}_{-0.5}$} & \textit{65.9$^{+0.7}_{-0.9}$} & \textit{71.2$^{+0.8}_{-0.7}$} \\
&ATT-TF-PERT & 88.4$^{+0.4}_{-0.3}$ & \textit{85.4$^{+0.9}_{-0.6}$ }& 77.2$^{+0.5}_{-0.4}$& 78.2$^{+0.4}_{-0.5}$ & 65.7$^{+0.5}_{-0.3}$& 61.3$^{+0.7}_{-0.5}$ & 65.3$^{+0.7}_{-0.8}$ & 67.4$^{+0.6}_{-0.8}$\\
\midrule
\multirow{8}{*}{\rotatebox[origin=c]{90}{sBERT}}& Sentence Average & 85.1$^{+0.6}_{-0.7}$ & 85.2$^{+0.6}_{-0.7}$ & 75.7$^{+0.6}_{-0.8}$ & \textit{78.2$^{+0.6}_{-0.7}$} & \textit{64.5$^{+0.7}_{-0.5}$} & 60.4$^{+0.8}_{-0.6}$ & \textit{66.4$^{+0.8}_{-0.7}$} & \textit{69.5$^{+0.8}_{-0.7}$}\\
& Top-Half Avg.  & 83.2$^{+0.8}_{-0.6}$ & 84.1$^{+0.7}_{-0.6}$ & 71.3$^{+0.5}_{-0.6}$& 76.5$^{+0.8}_{-0.6}$ & 60.8$^{+0.7}_{-0.9}$ & 60.4$^{+0.9}_{-1.2}$ & 62.8$^{+0.8}_{-0.7}$ & 63.5$^{+0.9}_{-0.8}$ \\
&Bottom-Half Avg. & 80.6$^{+0.7}_{-0.6}$& 81.3$^{+0.5}_{-0.8}$ & 66.5$^{+0.4}_{-0.4}$ & 70.1$^{+0.6}_{-0.4}$& 56.5$^{+0.4}_{-0.8}$& 58.7$^{+0.5}_{-0.6}$& 56.1$^{+0.7}_{-0.6}$ & 60.5$^{+0.5}_{-0.4}$\\
&TF-IDF Weighted &  84.2$^{+0.4}_{-0.5}$ & 82.8$^{+0.5}_{-0.4}$ & 75.1$^{+0.6}_{-0.7}$ & 74.3$^{+0.4}_{-0.6}$ & 63.2$^{+0.3}_{-0.2}$ & 61.2$^{+0.4}_{-0.5}$ & 63.4$^{+0.5}_{-0.3}$ & 65.8$^{+0.7}_{-0.6}$\\
&TK-PERT &  86.2$^{+0.6}_{-0.7}$ & 84.1$^{+0.8}_{-0.7}$ & 73.9$^{+0.6}_{-0.6}$& 77.1$^{+0.8}_{-0.6}$ & 62.6$^{+0.6}_{-0.8}$ & 59.3$^{+0.7}_{-0.5}$ & 65.4$^{+0.8}_{-0.6}$ & 68.1$^{+0.6}_{-0.7}$\\
&TF-PERT &  85.8$^{+0.5}_{-0.4}$ & 83.7$^{+0.2}_{-0.4}$& 72.7$^{+0.6}_{-0.5}$& 76.5$^{+0.4}_{-0.3}$ & 62.0$^{+0.6}_{-0.5}$& 60.4$^{+0.3}_{-0.6}$ & 64.3$^{+0.4}_{-0.8}$ & 68.2$^{+0.7}_{-0.8}$\\
&ATT-PERT &  \textit{88.5$^{+0.7}_{-0.6}$} & \textit{85.8$^{+0.5}_{-0.5}$ }& \textit{76.2$^{+0.8}_{-0.4}$}& 77.4$^{+0.5}_{-0.6}$ & 62.1$^{+0.6}_{-0.7}$& 60.8$^{+0.3}_{-0.6}$ & 66.1$^{+0.7}_{-0.4}$ & \textit{69.5$^{+0.8}_{-0.6}$}\\
&ATT-TF-PERT & 85.6$^{+0.5}_{-0.6}$ & 84.3$^{+0.3}_{-0.4}$ & 75.1$^{+0.6}_{-0.6}$& 76.8$^{+0.5}_{-0.6}$ & 61.3$^{+0.8}_{-0.5}$& \textit{62.7$^{+0.4}_{-0.5}$}& 65.8$^{+0.5}_{-0.3}$ & 66.4$^{+0.6}_{-0.6}$\\
\bottomrule
    \end{tabular}
    \caption{Accuracy for MLDoc classification on the zero-shot transfer task. Best results per language are shown in bold and per family in italics.}
    \label{tab:zero_mldoc}
\end{table*}


\paragraph{Multi-label ICD Code Classification} shows the same trend with respect to different sentence embeddings as above for German and Spanish, with a slight preference towards LaBSE embeddings. Table \ref{tab:icd-classification} shows the results for this task. There is a large discrepancy between the scores for the German and the Spanish datasets, as already noticed by the evaluations in the original corresponding shared tasks. The classification in Spanish achieves much lower results probably because of a very small training corpus.
Our results indicate that the information is spread throughout documents in this case. The difference between only using the top of the document and only using the bottom part is small, and using the whole document either by sentence averaging or considering it a single unit is always  better than any of its parts at a 95\% significance level. Semantic (TF-IDF) and positional (TK-PERT) information is less relevant. For the German task, either considering the full document as a whole (\textit{All tokens}) or averaging all the sentences gives the highest performance. For the Spanish task, even with a very low overall quality, learning specific weights for different parts of the document (ATT-PERT) boosts the quality. Comparing ATT-PERT with TK-PERT, we find that the trainable alternative performs better for all languages and base embeddings considered, however, the improvements are not statistically significant for all base embeddings in the case of German. 
In general, the windowing approaches that combine semantics with position (TF-PERT and ATT-TF-PERT) do not perform significantly better than the pure positional methods (TK-PERT and ATT-PERT). This can be explained by looking a concrete example. 
Figure~\ref{fig:scores} shows the distribution of weights across a document from the CANTEMIST health record corpus for 8 configurations based on LASER embeddings.
The example shows that the effect of the $tfidf$ component in ATT-TF-PERT (configuration 7) is equivalent to move weight mass from ATT-PERT (configuration 6) into TF-IDF (configuration 3).  When this happens, the result is a score in the middle of the way between ATT-PERT and TF-IDF.
In this document, a medical diagnostic evaluation is detailed and includes patient information, past diagnoses, family medical history, as well potential evolution of the disease. We observe that while the `Sentence average' configuration places largely equivalent weights on all the sentences, the TF-IDF weights place more emphasis on the beginning and end of the document which stores information about the patient and the evolution of the disease respectively. This behaviour is similar to the one exhibited by the PERT family of methods: the weight pattern observed for configurations 3-7 remain quite consistent but vary in their intensity. 



\paragraph{Cross-lingual Document Classification} 
data allows us to test the embedding methods on 8 languages (Table~\ref{tab:zero_mldoc}). The languages belong to three families, Indo-European (Germanic, Romance and Slavic), Japonic and Sino-Tibetan. All languages are high-resourced and 
included in our pre-trained sentence representation models. MLDoc documents are shorter than 1,000 tokens with an average length of 275 tokens for English and 562 for Chinese; the other languages stay in the middle. Given that length, the methods that use different 510-sized excerpts of the documents do not differ much as all the excerpts are ---for most of the documents--- the same.

Accuracies in Table \ref{tab:zero_mldoc} show that the documents convey slightly more meaning at the top part than at the bottom (\textit{Top-Half Avg.} vs \textit{Bottom-Half Avg.}). The sentence average is a very strong baseline and, for half of the languages (English, German, Russian and Chinese), this is statistically significantly better at 95\% confidence level than treating the document as a single unit with LASER. The TF-IDF version is worse than the simple sentence average except for Japanese- although not significantly so. Japanese has the lowest accuracy for all the languages and a high difference between the information at the top and the bottom of its documents. In general, position (TK-PERT) is more important than semantics (TF-IDF) and learning task-specific weights (ATT-PERT) further increases accuracy.
Additional experiments with TF-PERT and ATT-TF-PERT do not show statistically significant improvements over their counterparts TK-PERT and ATT-PERT, similarly to the trend observed in the previous tasks.
%
For English, Chinese and Spanish, we are further able to compare the performance of pre-trained large-input transformers. Longformer achieves 92.3\% of accuracy for English, which is 4.1\% better than the 88.7\% that LASER achieves in the \textit{All tokens} configuration and about 2\% better than the best performing architecture, the sentence average of LaBSE embeddings (90.9\%). However, the latter is not statistically significant at 95\% confidence level. 
The result is different for Chinese and Spanish. In both cases, considering all tokens with LASER and sentence average are better than Longformer, although the difference is not statistically significant for Spanish. This indicates that smaller amounts of training data can prevent native full document-level embeddings to be extended to languages other than English.

\section{Summary and Conclusions}
\label{sec:conclusion}

In this work, we studied effective methods for developing multilingual document-level representations. We used state-of-the-art sentence-level embeddings as basic units and systematically compare different pooling methods to evaluate these representations at the document level. We performed exhaustive evaluations across three sentence embeddings models, three tasks and eight languages. A key takeway from our work is that natively training on full documents i.e. with more compute is not necessarily the best option and instead a clever combination of sentences can be sufficient. 

Our experiments show that specific \textbf{base sentence embedding models} (LASER, LaBSE, sBERT)  do not impact the performance of the document-level embeddings much. We observe similar performance amongst them across all experiments. However, it is to be noted that we experiment with 
languages that while being morphologically distinct, are well resourced and covered by the three base sentence-embedding models. It would be interesting to explore how models behave when embeddings have a lower quality. For this, one would need to create
evaluation datasets at the document level for low-resourced languages but this is out of the scope of this work.

We observed that a simple sentence average is a very strong \textbf{pooling strategy}, specially for classification tasks. Positional and contextual information is more important than semantic information for the final performance as exemplified by the fact that PERT-based weightings perform better than TF-IDF's in all the tasks. When combining both, positional and semantic information, we do not observe statistically significant improvements with respect to only including positional information. For the classification tasks which include a learnable layer, we extend TK-PERT to ATT-PERT (and the semantic counterparts) and include global trainable attention on the positional information. This global attention is beneficial in all the cases.

The \textbf{type of document} is also relevant to chose the best method. Long documents might have the most crucial information stored in different parts. For instance, webpages have a majority of their information in the first half of the document as we observed in the document alignment task. In this case, the positional information significantly outperforms any model that does not take it into account. 


\section*{Limitations}
One of the main focal points of this work is multilinguality. In the presented approaches, the multilinguality of the resultant document embeddings depends solely on the language coverage and cross-lingual transfer ability of the pre-trained sentence embeddings used as basic units. Document-level representations are as robust to new languages and scripts as the base sentence embeddings are. Cross-lingual transfer is a perpendicular dimension not studied in this work.

We introduce ATT-PERT, a new learnable approach for the combination of sentence embeddings. This model is therefore  of use for tasks with a learning/fine-tuning phase but it is not intended for ready-to-use multilingual document-level embeddings in contrast to the existing pre-trained sentence-level counterparts.

\section*{Acknowledgements}
This work has been supported by the German Federal Ministry of Education and Research (BMBF) under funding code 01IW20010 (CORA4NLP).


\bibliography{anthology,custom}
\bibliographystyle{acl_natbib}

\end{document}